\documentclass[10pt,twocolumn,letterpaper]{article}

\usepackage[pagenumbers]{cvpr} 

\usepackage{booktabs}
\usepackage{multirow}
\usepackage{gensymb}

\definecolor{cvprblue}{rgb}{0.21,0.49,0.74}
\usepackage[pagebackref,breaklinks,colorlinks,allcolors=cvprblue]{hyperref}
\usepackage{hyperref}
\usepackage[accsupp]{axessibility}
\newcommand{\pseudopara}[1]{\vspace{1mm}\noindent\textbf{#1.}}

\def\paperID{****} 
\def\confName{CVPR}
\def\confYear{2025}

\title{Matrix3D: Large Photogrammetry Model All-in-One}

\author{
    Yuanxun Lu$^1$\thanks{Equal contribution.} \ \thanks{This project was performed during Yuanxun Lu's internship at Apple.} \quad Jingyang Zhang$^{2*}$ \quad Tian Fang$^2$ \quad Jean-Daniel Nahmias$^2$ \quad Yanghai Tsin$^2$ \\ Long Quan$^3$ \quad Xun Cao$^1$ \quad Yao Yao$^1$\thanks{Corresponding author.} \quad Shiwei Li$^2$
    \and
    $^1$Nanjing University\\
    {\tt\small luyuanxun@smail.nju.edu.cn, \{caoxun,yaoyao\}@nju.edu.cn}
    \and
    $^2$Apple\\
    {\tt\small \{jingyang\_zhang,fangtian,jnahmias,ytsin,shiwei\}@apple.com}
    \and
    $^3$The Hong Kong University of Science and Technology\\
    {\tt\small quan@cse.ust.hk}
}

\begin{document}

\twocolumn[{%
\renewcommand\twocolumn[1][]{#1}%
\maketitle
\vspace{-4mm}
\includegraphics[width=\linewidth]{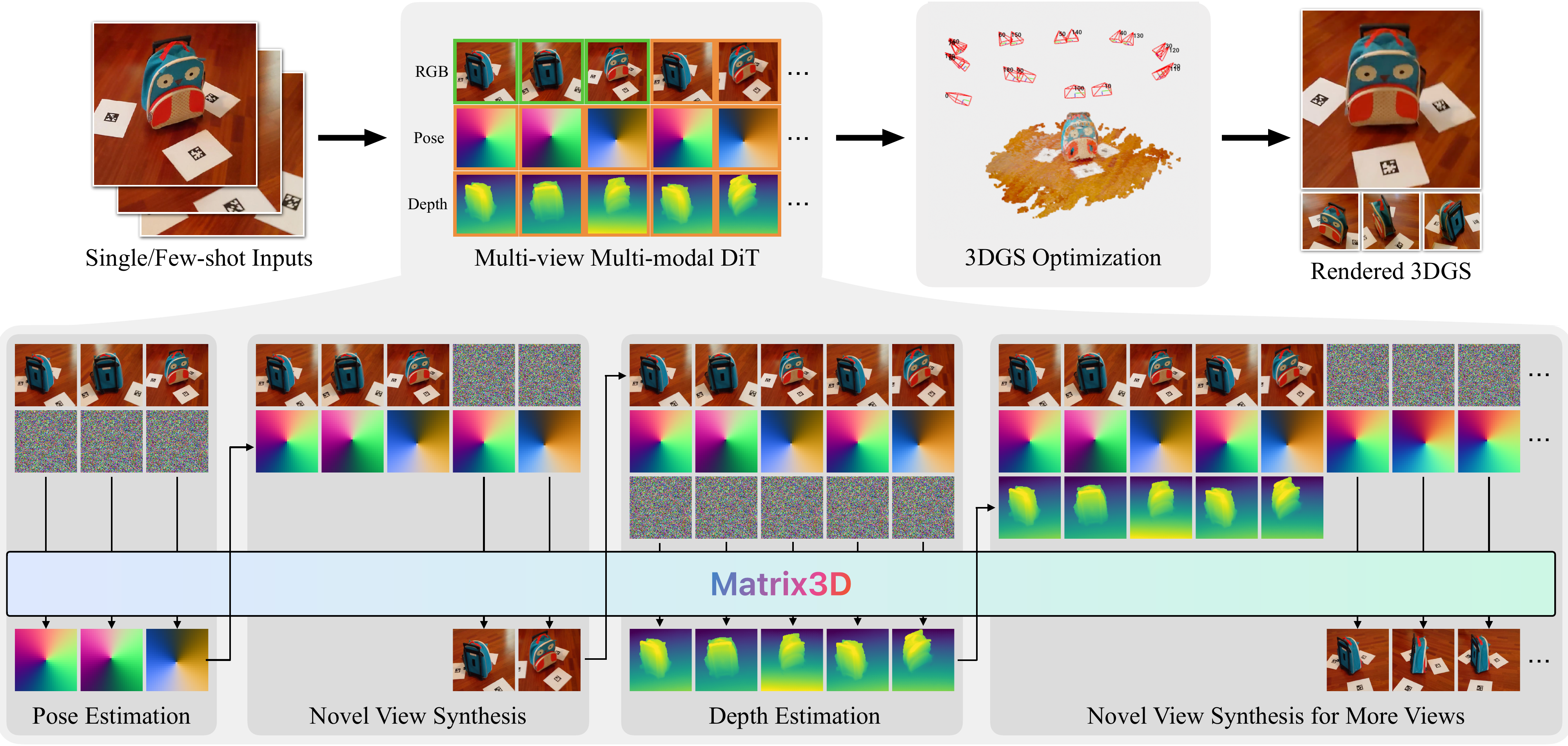}
\captionof{figure}{Utilizing \textbf{Matrix3D} for single/few-shot reconstruction. Before 3DGS optimization, we complete the input set by pose estimation, depth estimation and novel view synthesis, all of which are done by \textbf{the same} model. }
\label{fig:teaser}
\vspace{2mm}
}]

\renewcommand{\thefootnote}{\fnsymbol{footnote}}
\footnotetext[1]{Equal contribution.}
\footnotetext[2]{This project was performed during Yuanxun Lu's internship at Apple.}
\footnotetext[3]{Corresponding author.}

\begin{abstract}
We present Matrix3D, a unified model that performs several photogrammetry subtasks, including pose estimation, depth prediction, and novel view synthesis using just the same model. Matrix3D utilizes a multi-modal diffusion transformer (DiT) to integrate transformations across several modalities, such as images, camera parameters, and depth maps. 
The key to Matrix3D's large-scale multi-modal training lies in the incorporation of a mask learning strategy. This enables full-modality model training even with partially complete data, such as bi-modality data of image-pose and image-depth pairs, thus significantly increases the pool of available training data.
Matrix3D demonstrates state-of-the-art performance in pose estimation and novel view synthesis tasks. Additionally, it offers fine-grained control through multi-round interactions, making it an innovative tool for 3D content creation. Project page: \href{https://nju-3dv.github.io/projects/matrix3d}{https://nju-3dv.github.io/projects/matrix3d}.

\end{abstract}    
\section{Introduction}
\label{sec:intro}

Photogrammetry is a crucial technology for reconstructing 3D scenes from 2D images. However, the traditional photogrammetry pipeline has two significant weaknesses. First, it typically requires a dense collection of images--often hundreds--to achieve robust and accurate 3D reconstruction, which can be troublesome in practical applications. Second, the pipeline involves multiple processing stages that utilize completely different algorithm blocks, including feature detection, structure-from-motion (SfM), multi-view stereo (MVS), and etc. While each step is an independent task that requires unique algorithm, they are not correlated or jointly optimized with one another, which can result in suboptimal outcomes and accumulated errors throughout this multi-stage process.

The first challenge is commonly tackled by combining reconstruction with generation \cite{shi2023mvdream, li2023instant3d, wu2024reconfusion, gao2024cat3d}, where denser RGB images are generated by diffusion models conditioned on sparse inputs. In practice, however, when there are more than one input image, it is challenging to obtain accurate relative poses of them since they often come with low overlaps. The second challenge has not yet been extensively addressed. Representative works are PF-LRM \cite{wang2023pf} and DUSt3R \cite{wang2024dust3r}, which use single feed-forward models to perform both pose estimation and scene reconstruction. They are thus end-to-end optimizable and eliminate the need of multi-stage processing.

Inspired by previous methods, we try to take one step further and tackle these two challenges together, by building a unified model that can do multiple photogrammetry sub-tasks, including pose estimation, depth estimation and novel view synthesis (for sparse view reconstruction). We call this model \textbf{Matrix3D}, featuring an all-in-one generative model designed to support various sub-tasks in photogrammetry, through altering input/output combinations.
At its core, Matrix3D represents data of all modalities using unified 2D representations: camera geometries are encoded as Plücker ray maps, while 3D structures are presented as 2.5D depth maps. This makes it possible to leverage the capabilities of modern image generative models. 
We extend the diffusion transformer into a multi-view, multi-modal framework, capable of generating all necessary modalities. Inspired by the principles of masked auto-encoder (MAE), our model is trained by randomly masking inputs, while predicting the remaining unseen observations.
This masking learning design not only effectively manages varying degrees of input sparsity, but also substantially increases the volume of available training data by utilizing partially complete data samples such as bi-modality image-pose and image-depth pairs. 

With the densified camera / image / depth predictions generated by Matrix3D, a 3D Gaussian Splatting (3DGS)~\cite{kerbl20233d} optimization can be applied to produce the final output, where depth maps are back-projected into 3D point clouds for 3DGS initialization.

In summary, our key contribution is the unified diffusion transformer model that has flexible input/output configurations and enables several tasks in photogrammetry process. The proposed method eliminates the need for multiple task-specific models, streamlining the photogrammetry process with one single model. Extensive quantitative and qualitative experiments, demonstrate the state-of-the-art performance of our methods compared with existing task-specific approaches.
\section{Related Work}
\label{sec:related}

Photogrammetry, also known as image-based 3D reconstruction, is a foundational pillar in the field of 3D vision. A typical photogrammetry pipeline consists of several critical sub-steps. Below, we provide an overview of the most relevant works related to image-based 3D reconstruction.

\textbf{Structure-from-Motion (SfM)} is a classical approach for simultaneously recovering sparse 3D structures and estimating camera poses from multiple overlapping 2D images. These pipelines~\cite{hartley2003multiple, crandall2012sfm, schoenberger2016sfm, cui2017hsfm} typically begin with camera parameter estimation through feature matching~\cite{harris1988combined, lowe2004distinctive, rosten2006machine, bay2006surf} across images, followed by bundle adjustment to jointly optimize the 3D point cloud and camera poses. Recent advances have focused on enhancing the robustness of SfM through learning-based feature extractors~\cite{detone2018superpoint, dusmanu2019d2, revaud2019r2d2, luo2020aslfeat}, improved image-matching techniques~\cite{sun2021loftr, sarlin2020superglue, chen2022aspanformer, lindenberger2023lightglue}, and neural bundle adjustment~\cite{lin2021barf, lindenberger2021pixel, xiao2023level, wang2024vggsfm}. However, challenges still persist in sparse input scenarios, though gradually alleviated. Limited observations still introduce multi-view ambiguity and performance degradation.

\begin{figure*}[t]
    \centering
    \includegraphics[width=\linewidth]{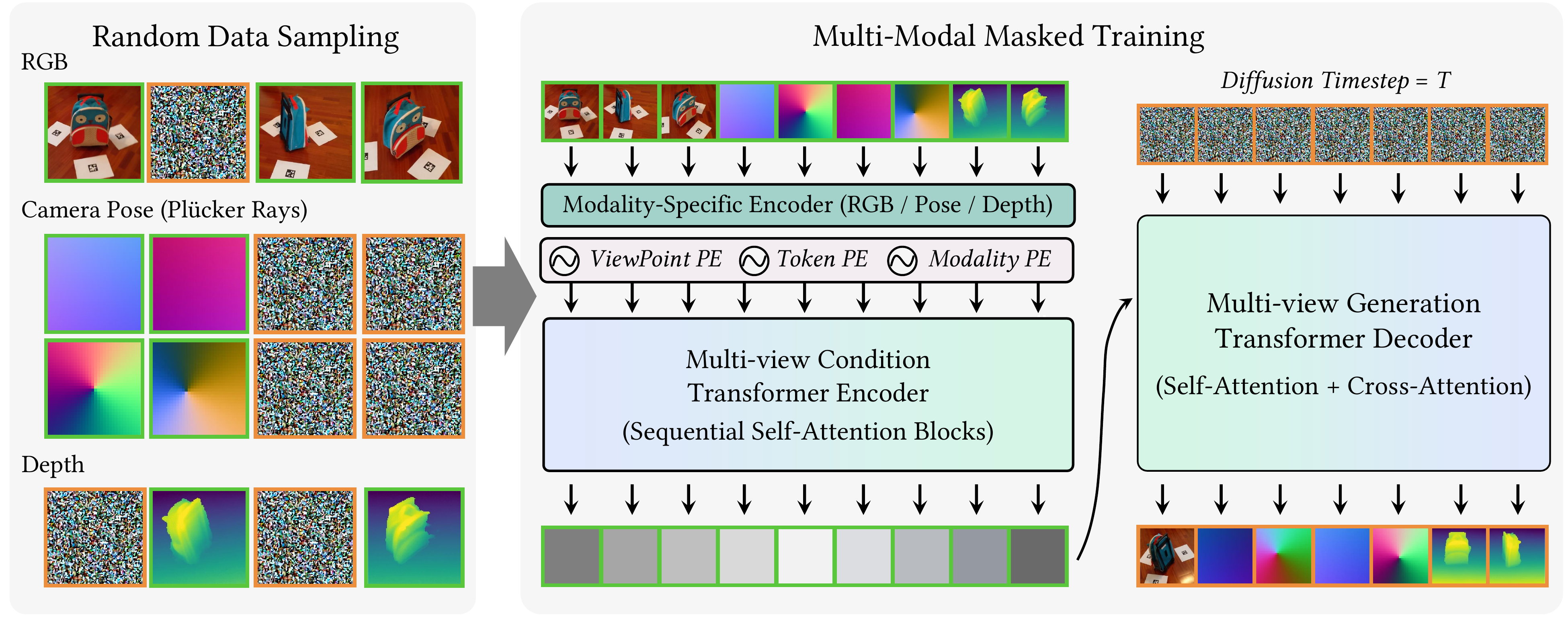}
    \vspace{-6mm}
    \caption{We train the Matrix3D by masked learning. Multi-modal data are randomly masked by noise corruption. Observations (green) and noisy maps (yellow) are fed into the encoder and the decoder respectively. By attaching the view and modality information to the clean and noisy inputs via different positional encodings, the model learns to denoise the corrupted maps and generate the desired outputs.}
    \vspace{-6mm}
    \label{fig:pipeline}
\end{figure*}

\textbf{Multi-view Stereo (MVS)} builds upon the camera poses obtained by SfM to create dense 3D geometry. Traditional MVS methods~\cite{hirschmuller2007stereo, schoenberger2016mvs, galliani2015massively, furukawa2015multi} depend on hand-crafted features and engineered regularizations to build dense correspondences and recover 3D points~\cite{lhuillier2005quasi, furukawa2009accurate}, volumes~\cite{kutulakos2000theory, seitz1999photorealistic, furukawa2009accurate}, or depth maps~\cite{campbell2008using, tola2012efficient, galliani2015massively}. Learning-based methods~\cite{yao2018mvsnet, yao2019recurrent, chen2019point, gu2020cascade, zhang2023vis} offer more powerful reconstruction with improved completeness and generalization capabilities. Typically, these methods assume well-calibrated camera parameters, which limits their robustness and applicability in real-world scenarios where calibration may be imprecise or unavailable.

\textbf{Sparse-view Pose Estimation} is extremely challenging due to the limited overlap between input images, making it difficult for traditional methods~\cite{snavely2006photo, mur2015orb} to build correspondences. Recent research has explored various strategies to address these ambiguities and predict relative poses from sparsely sampled images, including energy optimization~\cite{lin2023relpose++, sinha2023sparsepose}, exploiting synthetic data~\cite{jiang2024few}, leveraging data priors~\cite{wang2023posediffusion}, and applying probabilistic models~\cite{chen2021wide, zhang2022relpose}. Recently, RayDiffusion~\cite{zhang2024cameras} employs ray-based cameras and diffusion methods to model pose distributions, while PF-LRM~\cite{wang2023pf} and DUSt3R~\cite{wang2024dust3r} predict point maps in reference frames and recover poses using PnP algorithms.

\textbf{Feed-forward RGB-to-3D} approaches aim to directly infer 3D representations from single or a few RGB images via feed-forward models, without requiring per-scene optimization. These methods leverage strong 3D priors learned from large-scale object-level~\cite{chang2015shapenet} or larger datasets~\cite{deitke2023objaverse} to address inherent ambiguities. Pre-trained geometry models, such as monocular depth predictors~\cite{ranftl2021vision, bian2021auto, yin2021learning, Ranftl2022}, are usually adopted to further improve robustness. Recently, people proposed feed-forward methods~\cite{hong2023lrm, li2023instant3d, xu2023dmv3d, zhang2024gs, tang2024lgm, wu2024direct3d} which combines large transformer models to directly map RGB images into 3D representations~\cite{Chan2022Efficient, gao2024relightable, lin2024gaussian}. While these methods enable efficient 3D generation, their results remain less accurate than optimization-based methods.

\textbf{3D Generation with 2D Priors} refers to methods that use pre-trained 2D vision models to guide 3D generation. DreamFusion~\cite{poole2022dreamfusion} first proposed Score Distillation Sampling (SDS) to synthesize NeRFs from text by iteratively distilling knowledge from text-to-image diffusion models. Subsequent research has enhanced performance by improving distillation strategies~\cite{huang2023dreamtime, wang2023prolificdreamer, chen2023fantasia3d, zeng2024stag4d, haque2023instruct, tang2023dreamgaussian, zhang2024jointnet} or fine-tuning 2D diffusion models with camera conditioning~\cite{wu2024reconfusion, liu2023zero, raj2023dreambooth3d, chan2023generative, gu2023nerfdiff, sargent2023zeronvs, liu2024one}. Recent studies have shown that fine-tuning the model to generate multi-view images simultaneously~\cite{shi2023mvdream, tang2023mvdiffusion, liu2024one2345, shi2023zero123++, yang2024consistnet} provides stronger priors. Beyond SDS, researchers also explored offline 3D reconstruction~\cite{liu2023syncdreamer, long2023wonder3d, lu2024direct2, gao2024cat3d} directly from the generated multi-view information. Moreover, fine-tuning video diffusion models, which inherently encode 3D knowledge, has emerged as a promising approach~\cite{kwak2024vivid, voleti2024sv3d, melas20243d, guo2023animatediff}, though their higher computational demands remain a challenge.

\textbf{Masked Learning} has achieved significant success in pre-training tasks for NLP~\cite{kenton2019bert, brown2020language, radford2018improving, radford2019language} and computer vision~\cite{he2022masked, bachmann2022multimae, li2021mst}. These methods are proven to capture high-level semantics by masking parts of input and training models to reconstruct the masked contents. Recent research has extended this idea to multi-view image settings~\cite{weinzaepfel2022croco, weinzaepfel2023croco}, demonstrating improvements in downstream tasks like optical flow and stereo matching. In this work, we further apply masked learning to multi-view and multi-modal training to develop an all-in-one photogrammetry model.

\section{Method}
\label{sec:method}
In this section, we introduce Matrix3D, an all-in-one photogrammetry model for unified 3D reconstruction and generation. 
In the following, we describe the details of the framework design (Section~\ref{sec:3.1}), masking strategies (Section~\ref{sec:3.2}), dataset preparation (Section~\ref{sec:3.3}), training setup (Section~\ref{sec:3.4}), and downstream tasks (Section~\ref{sec:3.5}).

\subsection{Multi-Modal Diffusion Transformer}
\label{sec:3.1}
As demonstrated in Section~\ref{sec:intro}, our framework is designed around three key principles: a unified probabilistic model, flexible I/O, and multi-task capability. The emerging diffusion transformer (DiT)~\cite{peebles2023scalable} offers an ideal foundation, with its transformer architecture naturally supporting flexible I/O configurations and multi-modal fusion.

\pseudopara{Network architecture} 
The proposed model consists of two novel components compared with a standard image DiT model: a multi-view encoder and a multi-view decoder $D$. The encoder processes conditioning data from multiple views across different modalities (i.e., RGB, poses and depth), and embeds them into a shared latent space, enabling better cross-view and cross-modal feature integration. Similarly, the decoder processes noisy maps corresponding to different targets at different diffusion timestamps. The latent codes from different views and modalities are concatenated sequentially and passed through transformer layers to capture correspondence across views and modalities. Both the encoder and decoder are composed of multiple self-attention blocks, with the decoder additionally incorporating a cross-attention block after each self-attention layer to enable communication between conditioning inputs and generated outputs. Our diffusion model is built upon the pre-trained Hunyuan-DiT~\cite{li2024hunyuan} architecture with the aforementioned module modifications. In our experiments, the maximum number of views is set to 8, though this number can be further extended subject to computation budget. Mathematically, let $\mathbf{x}_c$ denotes multi-view/modality conditions, $\mathbf{x}_{g,t}$ the desired generation corrupted by noise $\epsilon$ at time $t$, and $\mathbf{x}_0$ the groundtruth. The diffusion model is trained using v-prediction~\cite{salimans2022progressive} loss:
\begin{align}
    \mathcal{L} &= \mathbb{E}_{\mathbf{x}_0, \boldsymbol{\epsilon}, t, y} \left[ \left\| D(E(\mathbf{x}_c), \mathbf{x}_{g,t}, t) - \mathbf{v} \right\|^2 \right], \\
    \mathbf{v} & = \alpha_t \boldsymbol{\epsilon} - \sigma_t \mathbf{x}_0.
\end{align}

\pseudopara{Multi-modality encoding} To handle multiple modalities, we apply modality-specific encoding methods before feeding the data into transformers. Specifically, the VAE from SDXL~\cite{podell2023sdxl} is used to encode RGB images into low-dimensional latent space. For camera poses, we follow RayDiffusion~\cite{zhang2024cameras} to represent cameras as Pl\"ucker ray maps, which naturally takes the form of image-like 2D data. For depth modality, our model adopts multi-view aligned depth (i.e., affine-invariant depth), which would be converted into disparities (i.e., the inverse of depth) to ensure a more compact data range. 
Additionally, a fixed shift and scale factor is applied to regularize ray maps and depth maps so that their distribution is closer to standard Gaussian distribution, as required by the diffusion process~\cite{ho2020denoising}.

\pseudopara{Positional encoding} We incorporate three types of positional encodings to preserve spatial relationships across viewpoints, patch token positions, and modalities. Specifically, we apply Rotary Positional Embedding (RoPE)~\cite{su2024roformer} to encode the positions of individual patch tokens because we care more about their relative position, while apply absolute sinusoidal positional encoding~\cite{dosovitskiy2020image} to viewpoints and modalities, each with different base frequencies, because we only need to distinguish between different view or modality IDs.

\subsection{Masked Learning}
\label{sec:3.2}

Unlike masked autoencoders (MAE), which mask portions of a single image, we extend this concept to the image level across multi-view, multi-modal settings, following a similar approach to 4M~\cite{mizrahi20244m}, to enable flexible I/O configurations. By masking specific views or modalities, the model learns to predict the missing content during both training and inference, facilitating dynamic and adaptable task handling.

\pseudopara{Training strategy} During training, in addition to the standard fully random masking strategy, we apply task-specific assignments. Specifically, we divide the training tasks into novel view synthesis, pose estimation, and depth prediction, along with the full random tasks, following a 3:3:3:1 ratio. We adopt a multi-stage training strategy: first training on 4-view models at 256 resolution, followed by 8-view models, and finally on 8-view models at 512 resolution.  Following Hunyuan-DiT~\cite{li2024hunyuan}, we use v-prediction~\cite{salimans2022progressive} as training objective. We adopt a 10\% probability of dropping conditions to enable classifier-free guidance (cfg)~\cite{ho2022classifier}. 

\subsection{Dataset Preparation}
\label{sec:3.3}
\pseudopara{Training data} We train Matrix3D on a mixture of six datasets: Objaverse~\cite{deitke2023objaverse}, MVImgNet~\cite{yu2023mvimgnet}, CO3D-v2~\cite{reizenstein21co3d}, RealEstate10k~\cite{zhou2018stereo}, Hypersim~\cite{roberts2021hypersim}, and ARKitScenes~\cite{baruch2021arkitscenes}. The first three datasets are indoor and object-centric, while the latter three cover large-scale indoor and outdoor scenes. 
Note that not all datasets provide all modalities and we use the available modalities for each dataset accordingly.

\begin{figure*}[t]
    \centering
    \includegraphics[width=\linewidth]{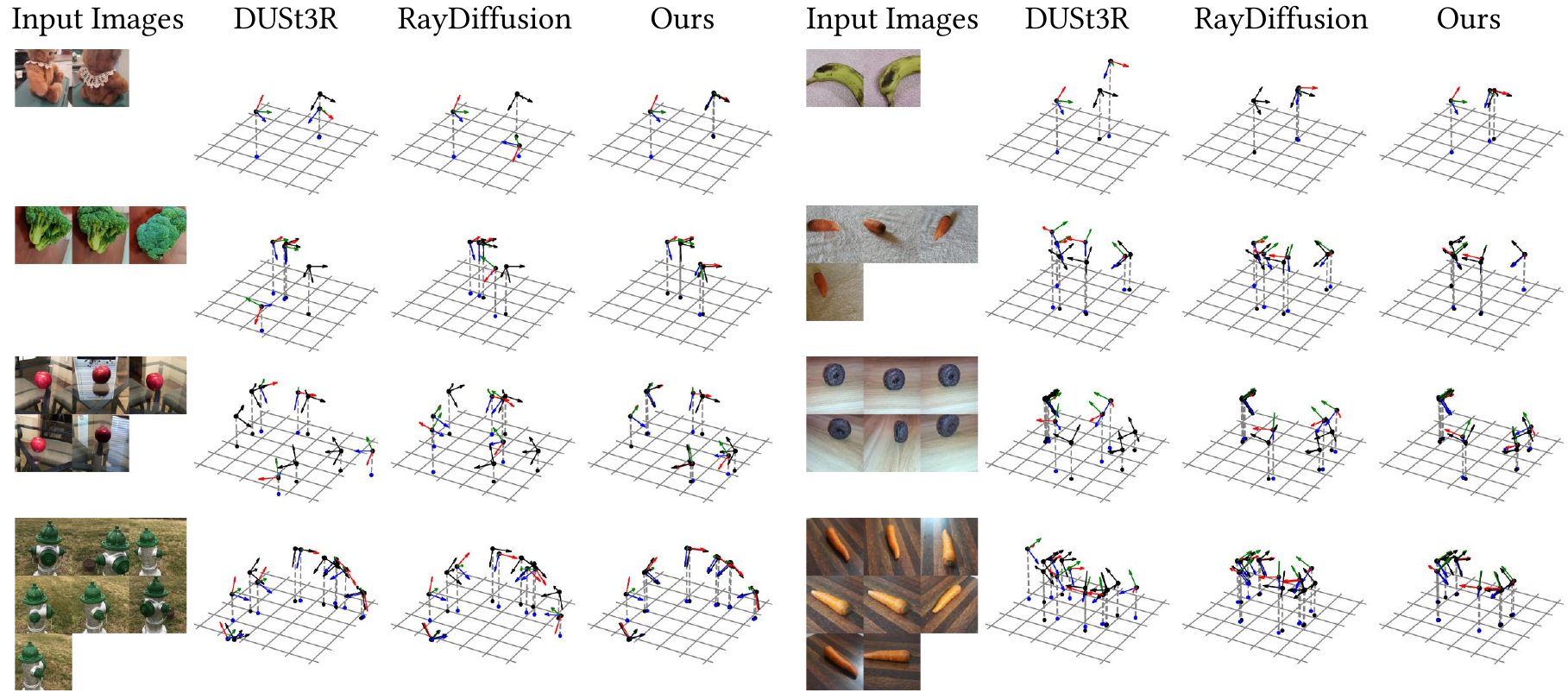}
    \vspace{-6mm}
    \caption{Sparse-view pose estimation results on CO3D dataset. The black axes are ground-truth and the colored ones are the estimation.
    }
    \vspace{-4mm}
    \label{fig:pose}
\end{figure*}

\begin{figure*}[t]
    \centering
    \includegraphics[width=\linewidth]{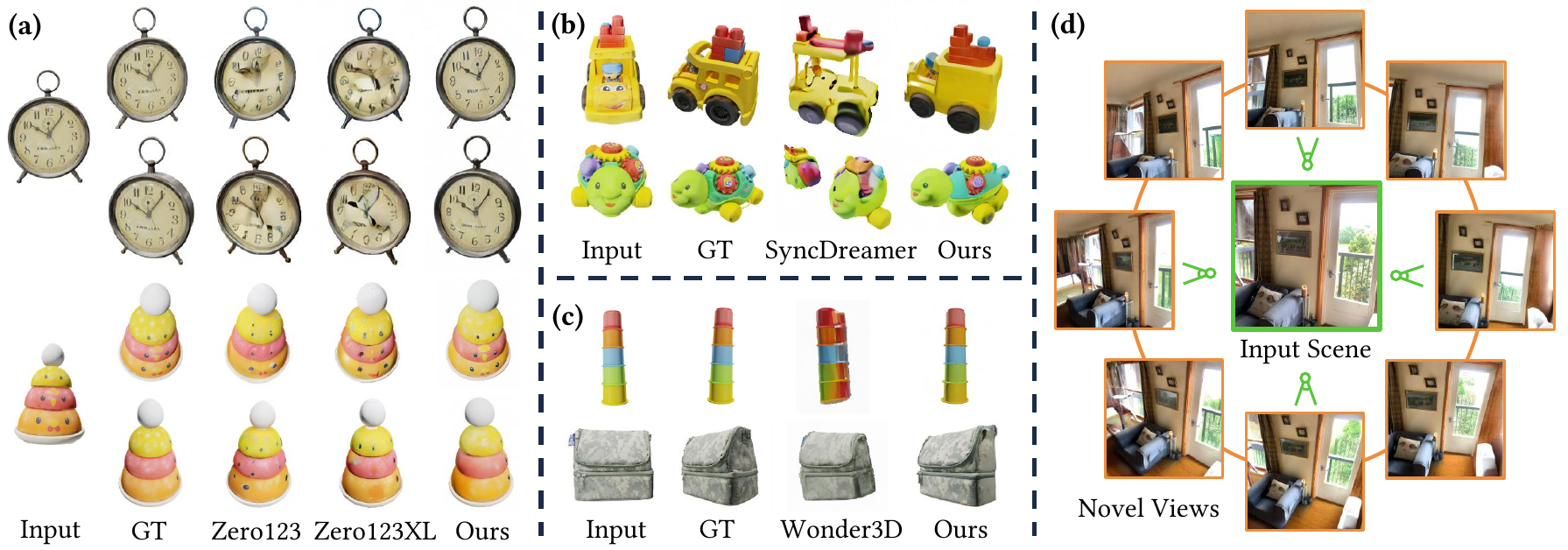}
    \vspace{-6mm}
    \caption{Qualitive evaluation results of novel view synthesis from single images on GSO and ARKitScenes dataset: a) random novel views; b) and c) follow the view configuration of SyncDreamer and Wonder3D respectively; d) indoor scenes from ARKitScenes dataset. Note that our method supports NVS of \textbf{arbitrary poses}.}
    \vspace{-4mm}
    \label{fig:nvs-gso}
\end{figure*}

\pseudopara{Normalization}  
Due to the highly diverse distributions of existing datasets, including variations in scale and scene type, preprocessing them consistently poses a challenge. To address this, we apply scene normalization and camera normalization. Please check the supplementary for details.



\pseudopara{Incomplete depth}
Real-world datasets often provide incomplete depth. A default solution to use these data is discarding patches with invalid pixels, which results in low data utilization. Instead, we concatenate valid masks to all the depth maps for both condition maps and noisy maps, and let the model learn how to discard invalid pixels. 
This also allow us to utilize sparse depth input during inference. 

\begin{table*}[t]
    \centering
    \resizebox{0.9\linewidth}{!}{
    \begin{tabular}{l|ccccccc|ccccccc}
\toprule
Metrics & \multicolumn{7}{c|}{Relative Rotation Accuracy @ 15$^\circ$($\uparrow, \%$)} & \multicolumn{7}{c}{Camera Center Accuracy @ 0.1 ($\uparrow, \%$)} \\
\midrule
\# of Images & 2 & 3 & 4 & 5 & 6 & 7 & 8 & 2 & 3 & 4 & 5 & 6 & 7 & 8\\
\midrule

COLMAP (SP+SG)~\citep{schoenberger2016sfm} & 31.3 &  29.0 &  27.3 &  27.6 &  28.0 &  29.3 &  31.9  & 100.0 &  34.8 &  24.1 &  19.1 &  16.0 &  15.0 &  15.7\\
PoseDiffusion~\citep{wang2023posediffusion} & 73.6 & 74.3 & 74.6 & 75.4 & 76.0 & 76.7 & 76.9 & 100.0 &  75.1 &  66.4 &  62.5 &  60.2 &  59.1 & 58.1\\
RelPose++~\citep{lin2023relpose++}  & 79.8 & 80.8 & 82.0 & 82.7 & 83.0 & 83.4 & 83.7 & 100.0 &  82.5 &  74.7 &  70.7 &  68.2 &  66.5 &  65.0\\
DUSt3R~\cite{wang2024dust3r} & 85.6 & 88.6 & 90.1 & 90.7 & 91.3 & 91.7 & 92.0 & 100.0 & 87.8 & 83.9 & 81.3 & 80.3 & 80.1 & 79.2\\
RayDiffusion~\cite{zhang2024cameras} & 90.4 & 91.2 & 91.5 & 91.9 & 92.1 & 92.3 & 92.4 & 100.0 & 93.1 & 88.9 & 86.0 & 84.1 & 82.8 & 81.9\\
\midrule
Ours {\fontsize{8}{16}\selectfont RGB Only} & \underline{95.6} & \underline{96.0} & \textbf{96.3} & \textbf{96.5} & \textbf{96.5} & \underline{96.3} & \underline{96.1} & 100.0 & \underline{93.5} & \underline{91.7} & \underline{90.6} & \underline{90.0} & \underline{89.1} & \underline{87.8}\\
Ours {\fontsize{8}{16}\selectfont RGB + Depth} & \textbf{95.8} & \textbf{96.3} & \underline{96.2} & \textbf{96.5} & \textbf{96.5} & \textbf{96.4} & \textbf{96.3} & 100.0 & \textbf{93.8} & \textbf{92.0} & \textbf{91.5} & \textbf{91.0} & \textbf{90.4} & \textbf{89.5} \\
\bottomrule
\end{tabular}
}
    \vspace{-3mm}
    \caption{Pose evaluation on CO3D. The percentage of relative rotations within 15 degrees and camera center errors within 10\% of the groundtruth scene scale are reported. The best results are in \textbf{bold} and the second bests are \underline{underlined}.
    }
    \vspace{-4mm}
    \label{tab:co3d_pose}
\end{table*}

\subsection{Training Setup}
\label{sec:3.4}
We initialize our model with pre-trained Hunyuan-DiT checkpoints~\cite{li2024hunyuan}. 
In the first stage, we train the model for 180K steps with a learning rate of 1e-4 on 64 NVIDIA A100 80G GPUs. In the latter two stages, we train the model for 20K steps for each stage with a learning rate of 1e-5 on 128 GPUs. 
The entire training takes approximately 20 days.

\subsection{Downstream Tasks}
\label{sec:3.5}

\pseudopara{Reconstruction tasks as modality conversions}
The Matrix3D model is able to perform multiple downstream tasks including pose estimation, multi-view depth estimation, novel view synthesis, or any hybrid of these tasks. By feeding any combination of conditional information and the desired output maps as noise, the model denoises the outputs according to the learned joint distribution. For example, pose estimation can be conducted by providing images of all input views, the identity camera pose for the reference view, and noisy ray maps for other views; novel view synthesis can be formulated as providing posed images for all reference views, ray maps for the novel views, and noisy novel-view images. 
Moreover, our model allows any reasonable input/output combinations, which cannot be achieved by previous task-specific models. For example, Matrix3D can achieve better results during NVS and pose estimation if depth maps are additionally provided. 

\pseudopara{3DGS Optimization}
In this section, we describe how to utilize our model for single or few-shot image reconstruction. We first use the Matrix3D model to complete multi-modality input (images/poses/depth maps) and also the image viewpoints. 
For few-shot inputs, we 1) estimate their camera poses, which can only be achieved by external methods in previous few-shot reconstruction systems; 2) estimate depth for the inputs as an initialization for 3DGS optimization; 3) synthesize novel views for stabilizing 3DGS optimization. 
For single input, we 1) synthesize more images to reach 8 key views with relatively large baselines to have an overall coverage of the target object or scene; 2) estimate depth maps for these key views; 3) synthesize novel views to interpolate the key images. 
Finally, we forward all the previous results to an open-source 3DGS reconstruction~\cite{nerfstudio} with several modifications. It is noteworthy that the 3DGS reconstruction is specially tailored to mitigate the multi-view inconsistency among the generated images. 

Please check the supplementary materials for more details about architectures, data pre-processing, training, and 3DGS optimization designs.

\section{Experiments}
In the following, we present the experiment results of different photogrammetry tasks.

\subsection{Pose Estimation}
We first evaluate our model for pose estimation under sparse views on the CO3D dataset. The proposed model is compared with multiple types of previous works: 1) traditional SfM: COLMAP~\cite{schoenberger2016sfm}; 2) discriminative neural networks: RelPose++~\cite{lin2023relpose++} and DUSt3R~\cite{wang2024dust3r}; 3) generative neural networks: PoseDiffusion~\cite{wang2023posediffusion} and RayDiffusion~\cite{zhang2024cameras}. 

We evaluate two metrics: relative rotation accuracy and camera center accuracy following RayDiffusion. For each scene, we estimate all 7 source views out of 8 views in one batch. The metrics are evaluated in a pair-wise manner. Our method outperforms other baselines by a significant margin (Table~\ref{tab:co3d_pose}). Figure~\ref{fig:pose} also presents a qualitative comparison between our predictions and ground truth poses. Our method performs better by a large margin than all baselines.

\begin{table}[t]
    \centering
    \resizebox{1.0\linewidth}{!}{
    \begin{tabular}{ll|ccc}
\toprule
View Settings & Methods & PSNR $\uparrow$ & SSIM $\uparrow$ & LPIPS $\downarrow$ \\
\midrule
\multirow{2}*{\shortstack{$\phi \in \{0\degree, 360\degree\}$ \\ $\theta =30\degree$}} & SyncDreamer~\cite{liu2023syncdreamer} & \underline{19.22} & \underline{0.82} & \textbf{0.16} \\
 & Ours & \textbf{20.45} & \textbf{0.86} & \textbf{0.16} \\
\midrule
\multirow{2}*{\shortstack{$\phi \in \{0\degree, 360\degree\}$ \\ $\theta =0\degree$}} & Wonder3D~\cite{long2023wonder3d} & \underline{13.28} & \underline{0.78} & \underline{0.33} \\
 & Ours & \textbf{18.97} & \textbf{0.86} & \textbf{0.18} \\
\midrule
\multirow{2}*{\shortstack{$\phi \in \{0\degree, 360\degree\}$ \\ $\theta \in \{30\degree, -20\degree\}$}} & InstantMesh~\cite{xu2024instantmesh} & 13.78 & 0.80 & 0.25 \\
 & Ours & \textbf{18.66} & \textbf{0.85} & \textbf{0.19} \\
\midrule
\multirow{4}*{\shortstack{$\phi \in \{0\degree, 360\degree\}$ \\ $\theta \in \{-90\degree, 90\degree\}$}} & Zero123~\cite{liu2023zero} & 17.56 & 0.80 & \underline{0.18} \\
& Zero123-XL~\cite{deitke2024objaverse} & 18.75 & 0.81 & \textbf{0.17} \\
& Ours & \underline{18.87} & \underline{0.85} & 0.21 \\
& Ours +Depth & \textbf{19.16} & \textbf{0.86} & 0.19 \\
\bottomrule
\end{tabular}
}
    \vspace{-3mm}
    \caption{Novel view synthesis (diffusion samples) evaluation on GSO. Results are grouped by different view settings because some methods uses fixed poses. Best results are in \textbf{bold} and the second best are \underline{underlined}. $\phi$ and $\theta$ denote azimuth and elevation angles.}
    \vspace{-3mm}
    \label{tab:gso_nvs}
\end{table}

\begin{table}[t]
    \resizebox{1.0\linewidth}{!}{
    \begin{tabular}{l|cccccc}
    \toprule
    Method & $\boldsymbol{\delta_{1}}$$\uparrow$ & $\boldsymbol{\delta_{2}}$$\uparrow$ & $\boldsymbol{\delta_{3}}$$\uparrow$ & AbsRel$\downarrow$ & log10$\downarrow$ & RMS$\downarrow$  \\
    \midrule
    Metric3D v2~\cite{yin2023metric3d, hu2024metric3d}  & 0.969   & 0.992    & 0.996  & 0.064   & 0.039    & 75.538   \\
    Depth Anything v2~\cite{depth_anything_v1, depth_anything_v2} & 0.950 & 0.992 & 0.997  & 0.077 & 0.045 & 85.188     \\
    \midrule
    Ours \ {\fontsize{8.5}{16}\selectfont CFG=1.5}  & 0.985 & 0.997 & 0.999 & \textbf{0.036} & 0.023 & 47.806   \\ 
    Ours \ {\fontsize{8.5}{16}\selectfont w/o CFG}  & \textbf{0.992} & \textbf{0.999}  & \textbf{1.000}  & 0.038 & \textbf{0.022} & \textbf{40.214}   \\ 
    \bottomrule
\end{tabular}
}
\vspace{-3mm}
\caption{Quantitative evaluation of monocular metric depth prediction tasks on DTU. The best results are in \textbf{bold}. }
\vspace{-6mm}
\label{tab:mono_depth}
\end{table}

\subsection{Novel View Synthesis}
In this section, we benchmark novel view synthesis task on diffusion samples on GSO~\cite{downs2022google} dataset against prior multi-view diffusion methods, including Zero123~\cite{liu2023zero}, Zero123XL~\cite{deitke2024objaverse}, SyncDreamer~\cite{liu2023syncdreamer}, Wonder3D~\cite{long2023wonder3d}, and InstantMesh~\cite{xu2024instantmesh}. 
For methods generating novel views at fixed camera poses, we follow their view configuration and use different rendered ground truths when comparing with them. 
For methods allowing arbitrary view synthesis, we render 32 random viewpoints for evaluation. 
PSNR, SSIM, and LPIPS are adopted as evaluation metrics. Figure~\ref{fig:nvs-gso} shows qualitative results on CO3D and ARKitScenes dataset. The quantitative results are shown in Table~\ref{tab:gso_nvs}. Our method achieves the best results for most of the metrics.

\subsection{Depth Prediction}\label{sec:exp-depth}
\pseudopara{Monocular Depth Prediction} We first evaluate our model on the monocular metric depth prediction task. Although our model is trained on at least two views, we found that it can still predict high-quality depth from single images. Table~\ref{tab:mono_depth} shows the comparison against Metric3D v2~\cite{hu2024metric3d} and Depth Anything v2~\cite{depth_anything_v2}. We adopt the IDR~\cite{yariv2020multiview} subset of the DTU dataset, and metrics following previous methods~\cite{hu2024metric3d} to evaluate all methods. All methods are not trained on the DTU dataset. Our method performs significantly better than the baselines. Qualitatively, we found that monodepth methods often produce distorted geometry which cannot be recovered by global linear alignment. This may be resulted from focal ambiguity issue and large domain gap between object-centric and open-scene data.

\begin{table}[t]
    \resizebox{1.0\linewidth}{!}{
    \begin{tabular}{ll|cccc|cc}
    \toprule
    \multicolumn{2}{l}{Methods} & Pose & Range & Int. & Align & rel$\downarrow$ & $\tau \uparrow$  \\
    \midrule
    \multirow{2}*{(a)} & COLMAP~\cite{schoenberger2016sfm, schoenberger2016mvs} & $\checkmark$ & $\times$ & $\checkmark$ & $\times$  & \textbf{0.7}   & \textbf{96.5}    \\
    & COLMAP Dense~\cite{schoenberger2016sfm, schoenberger2016mvs} & $\checkmark$ & $\times$ & $\checkmark$ & $\times$  & 20.8   & 69.3  \\
    \midrule
    \multirow{3}*{(b)} & MVSNet~\cite{yao2018mvsnet} & $\checkmark$ & $\checkmark$ & $\checkmark$ & $\times$  & \textbf{(1.8)} & (86.0) \\
    & MVSNet Inv. Depth~\cite{yao2018mvsnet} & $\checkmark$ & $\checkmark$ & $\checkmark$ & $\times$  & \textbf{(1.8)} & (86.7) \\
    & Vis-MVSNet~\cite{zhang2023vis} & $\checkmark$ & $\checkmark$ & $\checkmark$ & $\times$  & \textbf{(1.8)} & \textbf{(87.4)} \\
    & MVS2D DTU~\cite{yang2022mvs2d} & $\checkmark$ & $\checkmark$ & $\checkmark$ & $\times$  & (3.6) & (64.2) \\
    \midrule
    \multirow{4}*{(c)} & DeMon~\cite{ummenhofer2017demon} & $\times$ & $\times$ & $\checkmark$ & $\|\mathbf{t}\|$  & 21.8   & 16.6 \\ 
    & DeepV2D~\cite{teed2018deepv2d} & $\times$ & $\times$ & $\checkmark$ & med & 7.7 & 33.0 \\
    & DUSt3R 224~\cite{wang2024dust3r} & $\times$ & $\times$ & $\times$ & med  & \textbf{2.76}   & \textbf{77.32} \\
    & DUSt3R 512~\cite{wang2024dust3r} & $\times$ & $\times$ & $\times$ & med  & 3.52   & 69.33 \\
    \midrule
    \multirow{9}*{(d)} & DeMon~\cite{ummenhofer2017demon} & $\checkmark$ & $\times$ & $\checkmark$ & $\times$  & 23.7   & 11.5  \\
    & Deepv2D~\cite{teed2018deepv2d} & $\checkmark$ & $\times$ & $\checkmark$ & $\times$  & 9.2   & 27.4  \\
    & MVSNet~\cite{yao2018mvsnet} & $\checkmark$ & $\times$ & $\checkmark$ & $\times$  & (4429.1)  & (0.1) \\
    & MVSNet Inv. Depth~\cite{yao2018mvsnet} & $\checkmark$ & $\times$ & $\checkmark$ & $\times$  & (28.7)  & (48.9) \\
    & Vis-MVSNet~\cite{zhang2023vis} & $\checkmark$ & $\times$ & $\checkmark$ & $\times$  & (374.2)  & (1.7) \\
    & MVS2D DTU~\cite{yang2022mvs2d} & $\checkmark$ & $\times$ & $\checkmark$ & $\times$  & (1.6) & (92.3) \\
    & Robust MVD Baseline~\cite{schroppel2022benchmark} & $\checkmark$ & $\times$ & $\checkmark$ & $\times$  & 2.7 & 82.0 \\
    & Ours & $\checkmark$ & $\times$ & $\checkmark$ & $\times$  & \textbf{1.83}   & \textbf{85.45} \\
    \bottomrule
\end{tabular}
}
\vspace{-3mm}
\caption{Quantitative evaluation of multi-view metric depth prediction tasks on DTU. The methods are categorized into: a) traditional methods; b) with poses and depth range; c) without poses and depth range, but with alignment; and d) with poses, without range and alignment. Parentheses
refers to training on the same set. The best results are in \textbf{bold}. Results of other methods are reported in DUSt3R~\cite{wang2024dust3r}.}
\vspace{-5mm}
\label{tab:multiview_depth}
\end{table}

\begin{figure}[t]
    \centering
    \includegraphics[width=\linewidth]{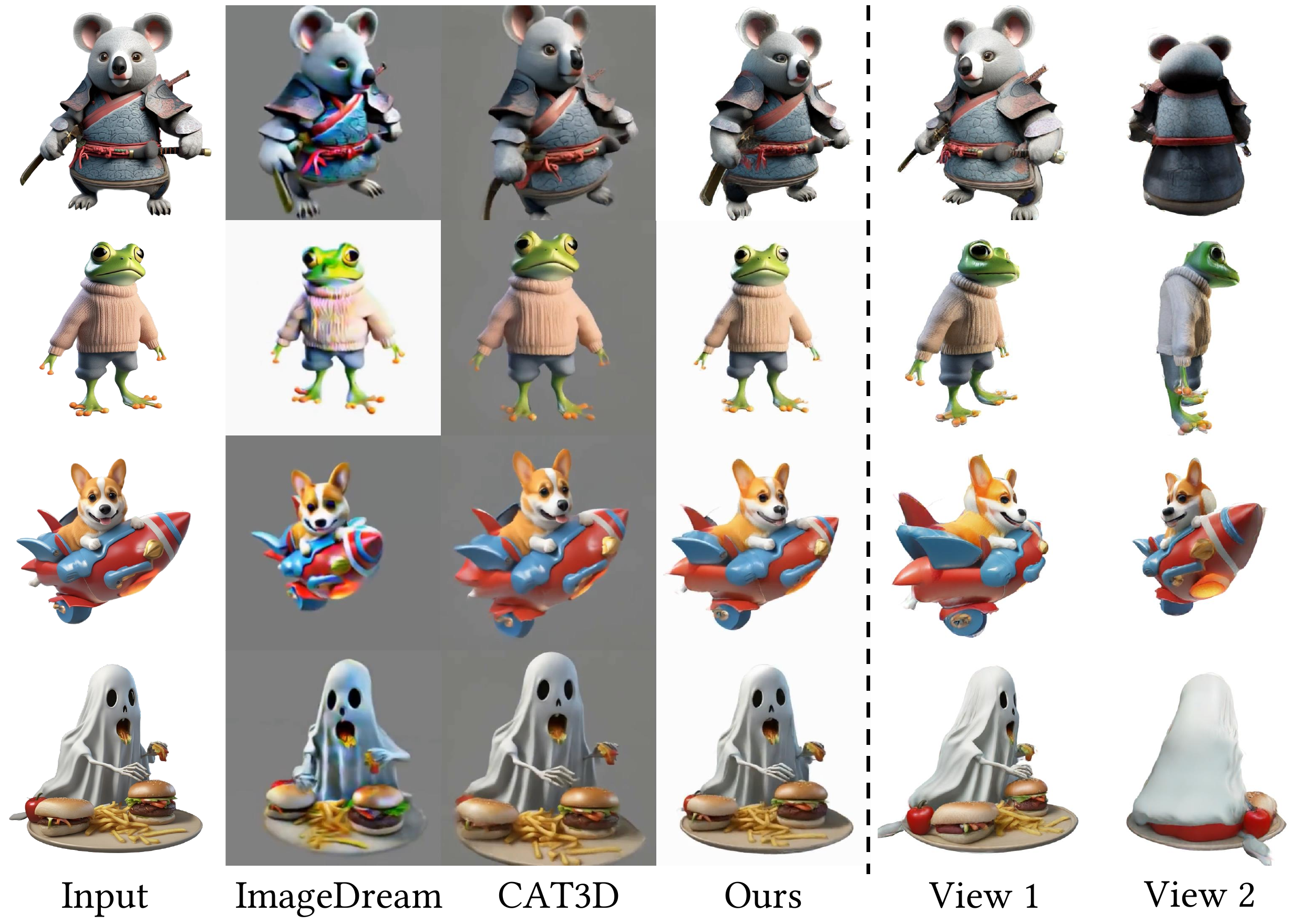}
    \vspace{-7mm}
    \caption{Monocular 3D reconstruction. Additional novel view renderings of our method are shown in the last two columns.
    }
    \vspace{-6mm}
    \label{fig:mono-recon}
\end{figure}

\pseudopara{Multi-view Depth Prediction} We then evaluate Matrix3D on the multi-view stereo depth prediction task. We evaluate it on the DTU~\cite{aanaes2016large} dataset. Following DUSt3R, we compute the Absolute Relative Error (rel) and Inlier Ratio on all test sets. Table~\ref{tab:multiview_depth} shows the quantitative results. We also back-project the depth maps to point clouds (with and without pose input) and evaluate their Chamfer distance to the ground truths. 
Quantitative results on DTU are shown in Table~\ref{tab:multiview_pcd}. More details about the conversion from depth maps to point clouds can be found in the supplementary material. 

Here we mainly discuss DUSt3R and our method which regress 3D information from images by a network architecture without any 3D-specific operations. We found that the results of our method is better than DUSt3R for depth maps, but is worst for point cloud. Given that our method also achieves higher pose estimation accuracy, one possible reason is that DUSt3R is supervised directly by point positions. So it can achieve good point cloud evaluation, but fails for the two decoupled tasks. Overall, these two methods cannot achieve the same accuracy level as the methods with 3D domain knowledge embedded. But the results are accurate enough to serve as good 3DGS initialization.

\begin{table}[t]
    \resizebox{1.0\linewidth}{!}{
    \begin{tabular}{ll|c|ccc}
    \toprule
    & Methods & GT cams & Acc.$\downarrow$ & Comp.$\downarrow$ & Overall$\downarrow$       \\
    \midrule
    \multirow{4}{*}{(a)} & Camp~\cite{campbell2008using} & $\checkmark$ & 0.835 & 0.554 & 0.695 \\
    &Furu~\cite{furukawa2009accurate} & $\checkmark$ & 0.613 & 0.941 & 0.777 \\
    &Tola~\cite{tola2012efficient} & $\checkmark$ & 0.342 & 1.190 & 0.766 \\
    &Gipuma~\cite{galliani2015massively} & $\checkmark$ &\textbf{ 0.283} & 0.873 & 0.578\\
    \midrule
    \multirow{4}{*}{(b)} &MVSNet~\cite{yao2018mvsnet} & $\checkmark$ &0.396 & 0.527 & 0.462 \\
    & CVP-MVSNet~\cite{Yang_2020_CVPR} & $\checkmark$ & 0.296 & 0.406 & 0.351 \\
    & UCS-Net~\cite{cheng2020deep} & $\checkmark$ & 0.338 & 0.349 & 0.344 \\
    & CIDER~\cite{xu2020learning} & $\checkmark$ & 0.417 & 0.437 & 0.427 \\
    & CasMVSNet~\cite{gu2020cascade} & $\checkmark$ & 0.325 & 0.385 & 0.355 \\
    & PatchmatchNet~\cite{wang2020patchmatchnet} & $\checkmark$ & 0.427 & 0.277 & 0.352 \\
    & Vis-MVSNet~\cite{zhang2023vis} & $\checkmark$ & 0.369 & 0.361 & 0.365 \\
    & CER-MVS~\cite{ma2022multiview} & $\checkmark$ & 0.359 & 0.305 & 0.332 \\
    & GeoMVSNet~\cite{zhe2023geomvsnet} & $\checkmark$ & 0.331 & \textbf{0.259} & \textbf{0.295} \\
    \midrule
    & DUSt3R~\cite{wang2024dust3r} & $\times$ &   2.677  &  0.805  & 1.741  \\ %
    & Ours & $\times$ &   3.261 & 1.170 & 2.266 \\ %
    & Ours & $\checkmark$ &   2.930 & 1.265 & 2.098 \\ %
    \bottomrule
    \end{tabular}
}
\vspace{-3mm}
\caption{Quantitative evaluation of point clouds back-projected from multi-view metric depths on DTU. The methods are categorized into: a) traditional methods and b) learning-based MVS. The best results are in \textbf{bold}. Results of other methods are reported in DUSt3R~\cite{wang2024dust3r}.}
\label{tab:multiview_pcd}
\vspace{-4mm}
\end{table}

\begin{table}[t]
    \resizebox{1.0\linewidth}{!}{
    \begin{tabular}{l|c|}
    \toprule
    Method & CLIP (Image) $\uparrow$\\
    \midrule
    ImageDream~\cite{wang2023imagedream} & 83.77 $\pm$ 5.2 \\
    One2345++~\cite{liu2024one2345} & 83.78 $\pm$ 6.4\\
    IM3D~\cite{melas20243d} & \textbf{91.40} $\pm$ 5.5\\
    CAT3D~\cite{gao2024cat3d} & 88.54 $\pm$ 8.6\\
    Ours & \underline{88.76} $\pm$ 7.2 \\
    \bottomrule
\end{tabular}
    \begin{tabular}{|l|ccc}
    \toprule
    Method & PSNR $\uparrow$ & SSIM $\uparrow$ & LPIPS $\downarrow$ \\
    \midrule
    Zip-NeRF~\cite{barron2023zip} & 14.34 & 0.496 & 0.652 \\
    ZeroNVS~\cite{sargent2023zeronvs} & 17.13 & 0.581 & 0.566 \\
    ReconFusion~\cite{wu2024reconfusion} & 19.59 & \underline{0.662} & 0.398 \\
    CAT3D~\cite{gao2024cat3d} & \textbf{20.57} & \textbf{0.666} & \textbf{0.351} \\
    Ours & \underline{20.02} & 0.633 & \underline{0.396} \\
    \bottomrule
\end{tabular}
}
\vspace{-3mm}
\caption{Quantitative evaluation of monocular (left) and posed few-shot (right) 3d reconstruction. The best results are in \textbf{bold} and the second bests are \underline{underlined}.}
\vspace{-3mm}
\label{tab:recon}
\end{table}

\begin{figure}[t]
    \centering
    \includegraphics[width=1.0\linewidth]{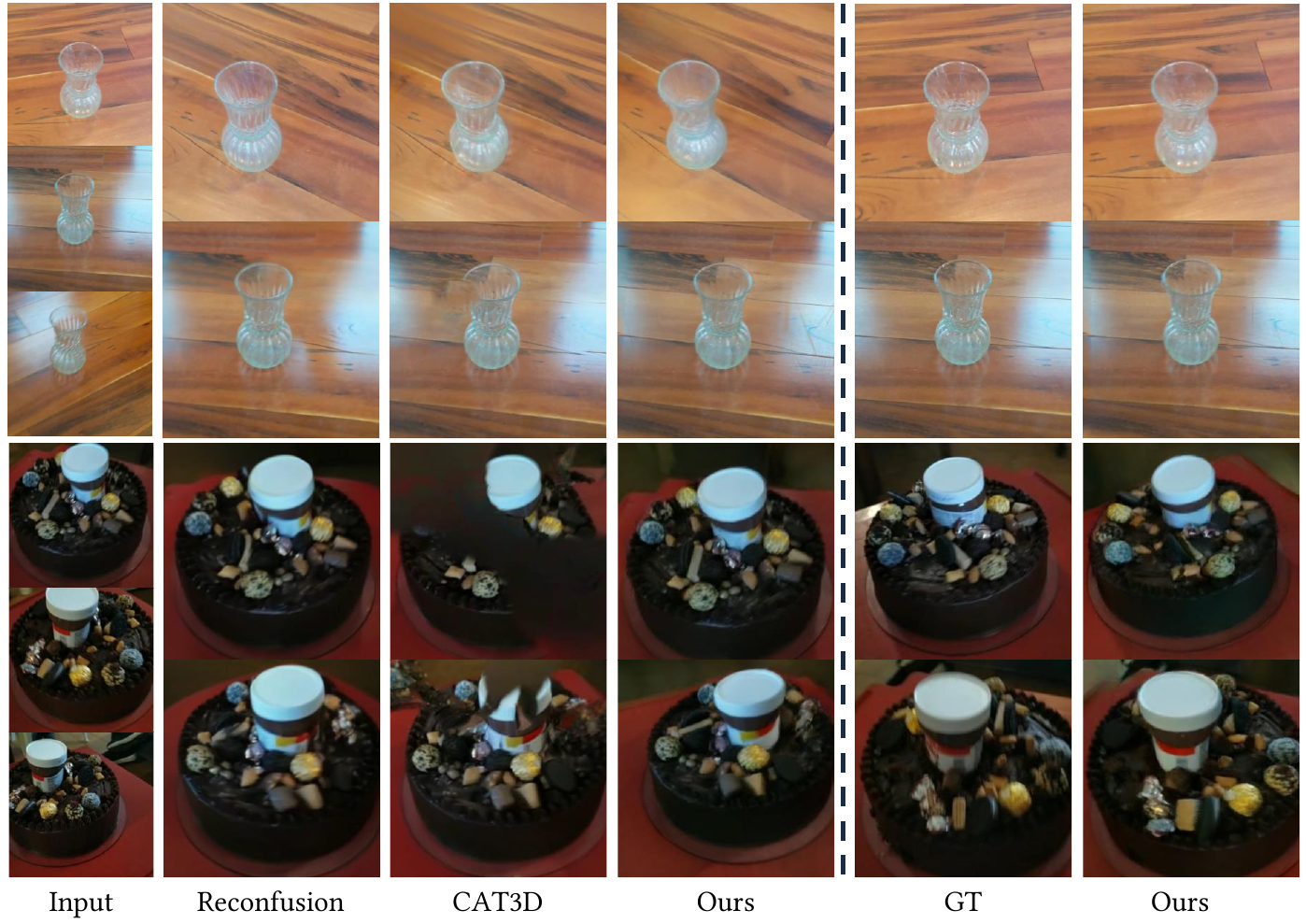}
    \vspace{-7mm}
    \caption{Sparse view 3D Gaussian Splatting reconstruction results from 3-view images input on CO3D dataset.}
    \vspace{-6mm}
    \label{fig:sparse-recon}
\end{figure}

\begin{figure*}[t]
    \centering
    \includegraphics[width=0.92\linewidth]{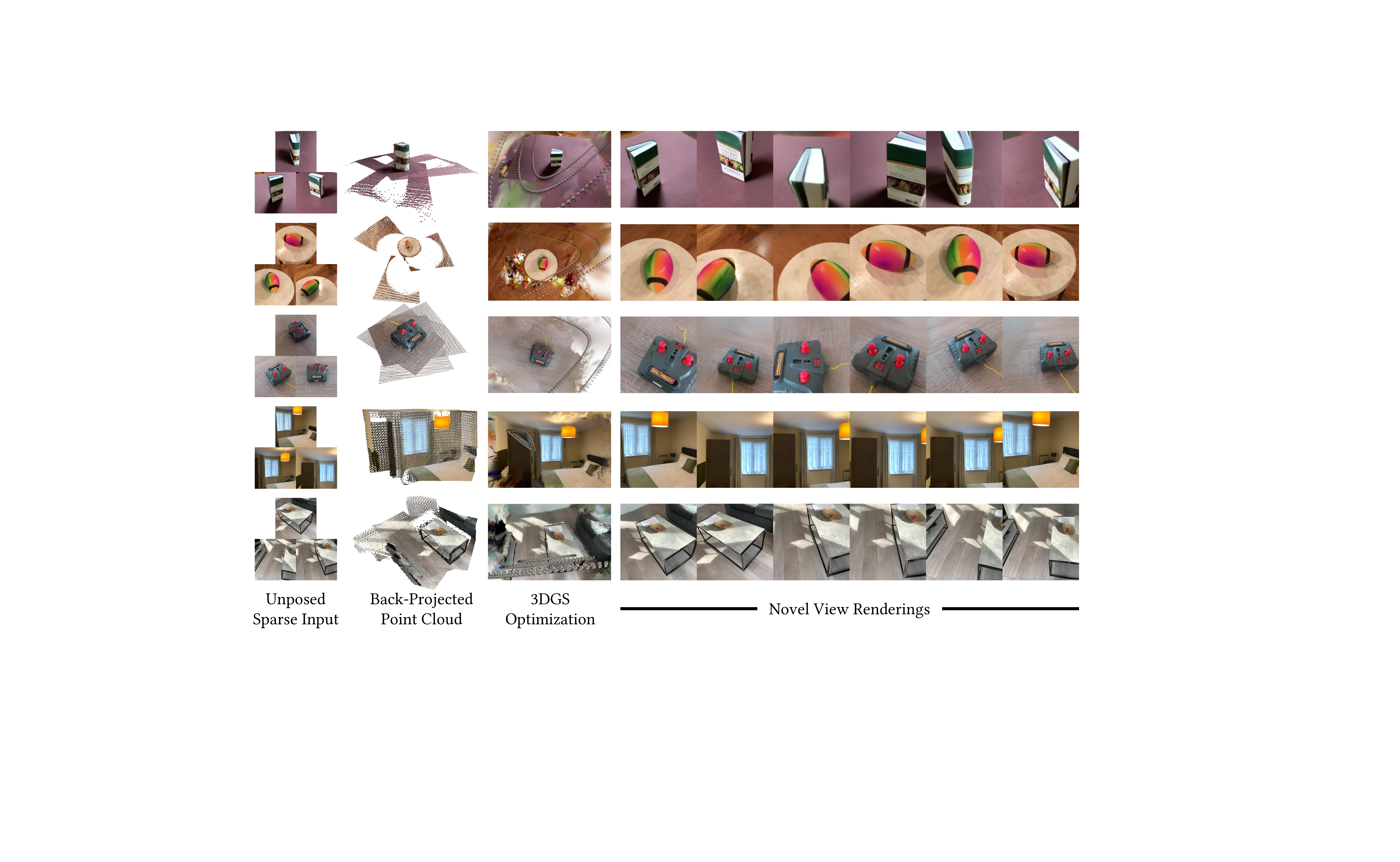}
    \vspace{-3mm}
    \caption{Unposed sparse-view 3D reconstruction results. }
    \vspace{-5mm}
    \label{fig:unposed-vis}
\end{figure*}

\subsection{3D Reconstruction}
\pseudopara{Monocular}
In the following, we evaluate the 3D reconstruction performance from single images. Specifically, we compare Matrix3D with diffusion-based optimization methods including ImageDream~\cite{wang2023imagedream}, One2345++~\cite{liu2024one2345}, IM-3D~\cite{melas20243d}, and CAT3D~\cite{gao2024cat3d} in terms of CLIP scores following CAT3D. Figure~\ref{fig:mono-recon} and Table~\ref{tab:recon} illustrate the comparisons. Our method achieves comparable results to SOTA methods.

\pseudopara{Sparse-view}
We perform 3D reconstruction from sparse-view unposed images. 
Although previous methods have made similar attempts, they require given poses which is challenging to estimate for traditional SfM methods in the case of sparse view, and just use ground truth for experiments, leaving pose estimation problem unsettled. 
Our framework seamlessly integrates the pose estimation and reconstruction process into a single pipeline. Figure~\ref{fig:unposed-vis} shows the results of unposed 3-view images from CO3D and ARKitScene datasets. The reconstruction process can be found in Sec.~\ref{sec:3.5}. Results show that our method successfully performs the reconstruction given unposed images.

We also evaluate the task using ground truth poses as input. Following previous methods, we conduct 3-view reconstruction experiments on the same train/test split of the CO3D dataset, as done in CAT3D\cite{gao2024cat3d}. We use PSNR, SSIM, and LPIPS to evaluate performance (Table \ref{tab:recon}). Figure~\ref{fig:sparse-recon} shows the comparisons. Note that we use fewer than half of the novel views employed in CAT3D for reconstruction. Our method performs slightly worse than CAT3D, primarily due to the smaller number of views used for 3DGS training. Additionally, while our reconstruction method can be run on a single GTX 3090 GPU, CAT3D requires 16 A100 GPUs, making it impractical for most users.

\subsection{Hybrid Tasks}\label{sec:exp-hybrid}
One major advantage of multi-modal masked learning is that the model can accept flexible input combinations. If additional information is provided, the model knows how to take advantage of them, and thus produces better outputs. We show this quantitatively by adding depth ground truth to the NVS and the pose estimation tasks mentioned above. For pose estimation, as shown by Table~\ref{tab:co3d_pose}, \textit{Ours RGB+Depth} consistently outperforms \textit{Ours RGB Only} in terms of camera center accuracy. One possible reason is that depth information mitigates scale ambiguity. For NVS, as shown by Table~\ref{tab:gso_nvs}, \textit{Ours+Depth} also achieves better results than \textit{Ours}. Intuitively, depth maps provide parital geometry, and thus facilitate NVS. In application, users can utilize depth measurement from active sensors to boost the performance of these two tasks.

\section{Conclusion}
In this paper, we introduced Matrix3D, a unified model that effectively addresses multiple photogrammetry tasks including pose estimation, depth prediction, and novel view synthesis using a multi-modal diffusion transformer (DiT). By employing a mask learning strategy, Matrix3D supports flexible input/output combinations, and maximizes training data from incomplete datasets. Through multi-round, multi-view, multi-model interactive generation, users can perform single or few-shot generation with one single model. Extensive experiments show that Matrix3D achieves SOTA performance in pose estimation and novel view synthesis tasks, showing its versatility on photogrammetry applications. 

\section{Acknowledgments}
This work is supported by National Natural Science Foundation of China (62472213, 62025108), Gusu Innovation \& Entrepreneurship Leading Talents Program (ZXL2024361), and Hong Kong RGC GRF 16206722.

{
    \small
    \bibliographystyle{ieeenat_fullname}
    \bibliography{main}
}


\clearpage
\setcounter{page}{1}
\maketitlesupplementary
Here, we present an additional description of the model architecture(Sec.~\ref{sec:supp-arch}), dataset preprocessing(Sec.~\ref{sec:supp-dataset}), training details(Sec.~\ref{sec:supp-training}), and experiments(Sec.~\ref{sec:supp-exp}).

\section{Model Architecture}
\label{sec:supp-arch}
For RGB data, we use DINOv2~\cite{oquab2023dinov2} and Stable Diffusion~\cite{rombach2021highresolution} VAE to extract deep features from pixels before sending them into the modality-specific encoders. The modality-specific encoders are composed of stacked convolution and linear layers following~\cite{peebles2023scalable} to patchify image-like 2D data into 1D tokens. The patchify scale for RGB, pose, and depth are set to 2, 1, and 4. After the tokens are processed by the transformer, we use similar modality-specific linear layers~\cite{peebles2023scalable} to unpatchify each modality token back to the original shape according to the corresponding patchify scales. The whole multi-view transformer encoder includes 20 self-attention blocks with a hidden size of 1024, while the decoder includes 40 stacked self-attention and cross-attention blocks with a hidden size of 1408 following HunyuanDiT~\cite{li2024hunyuan}. 

For classifier-free guidance (cfg), we empirically found the following settings to perform best: 1.5 for RGB / poses, and 1.0 for depth (w/o cfg).

\section{Dataset Pre-processing}
\label{sec:supp-dataset}
As illustrated in the main paper, we train Matrix3D on a mixture of six datasets, including Objaverse~\cite{deitke2023objaverse}, MVImgNet~\cite{yu2023mvimgnet}, CO3D-v2~\cite{reizenstein21co3d}, RealEstate10k~\cite{zhou2018stereo}, Hypersim~\cite{roberts2021hypersim}, and ARKitScenes~\cite{baruch2021arkitscenes}. In each training batch, the datasets have a proportion of 4:4:4:4:4:1. Table~\ref{tab:supp-datasets} provides a summary of these datasets used for training, including the size (in terms of scenes and images), type (real or synthetic), scene categories, and supported modalities (RGB, camera poses, and depths). For all datasets, we apply scene normalization and camera normalization. Camera poses are represented as Pl\"ucker rays. Note that the depth images provided in each dataset are not always complete. Specifically, CO3D-V2 and ARKitScenes provide incomplete depth images, while for the Objaverse dataset we only have the rendered object foreground depth.

\pseudopara{Normalization}  
Due to the highly diverse distributions of existing datasets, including variations in scale and scene type, preprocessing them consistently poses a challenge. To address this, we apply the following normalization.
\begin{itemize}
    \item Scene Normalization:  
    To normalize the whole scene scale, we adapt our approach depending on the dataset type and available modalities. For object-centric datasets with camera poses provided (i.e., Objaverse~\cite{deitke2023objaverse}, MVImgNet~\cite{yu2023mvimgnet}, and CO3D-v2~\cite{reizenstein21co3d}), we follow RayDiffusion~\cite{zhang2024cameras} by setting the intersection point of the input camera rays as the origin of the world coordinates and defining the scene scale accordingly. For scene-type datasets that provide depth information (i.e., Hypersim~\cite{roberts2021hypersim} and ARKitScenes~\cite{baruch2021arkitscenes}), we use the depth of the first view as a reference, calculating its median value and normalizing it to 1.0. For those datasets without depth data (i.e., RealEstate10k~\cite{zhou2018stereo}), we determine the scale based on the camera distances to the average positions and set the maximum distance to 1.0.

    \item Camera Normalization:  
    We perform camera normalization after scene normalization. Specifically, we set the first view’s camera as the identity camera with rotation \( R = I \) and translation \( T = [0, 0, 1] \), while preserving relative transformations between cameras across views.
\end{itemize}

\pseudopara{Objaverse Rendering}  For the Objaverse dataset, we render all models into RGB and depth images for training. Specifically, each 3D object is first normalized at the world center within a bounding box of $[-1, 1]^3$, and we render the whole scene from 32 random viewpoints. The render camera FoV is set to 50$\degree$. The azimuth and elevation angles are randomly sampled in $[0\degree, 360\degree]$ and $[-45\degree, 90\degree]$. The camera distance to the world center is randomly sampled in [1.1, 1.6], and the height on the z-axis is set in [-0.4, 1.2]. We use a composition of random lighting from area lighting, sun lighting, point lighting, and spot lighting.

\section{Training Details}
\label{sec:supp-training}
Table~\ref{tab:supp-hyperparams} reports the detailed training hyper-parameter settings of three stages. We didn't apply any data augmentation techniques and center-cropped the input images into a square.

\begin{table*}[]
    \centering
    \resizebox{1.0\linewidth}{!}{
    \begin{tabular}{l@{\hskip 1.0cm}l@{\hskip 0.8cm}l@{\hskip 0.8cm}l@{\hskip 0.8cm}l}
    \toprule 
    Hyper-parameters & Ablation & Stage 1 & Stage 2 & Stage 3 \\
    \midrule
    Optimizer & AdamW~\cite{loshchilov2017decoupled} & AdamW~\cite{loshchilov2017decoupled} & AdamW~\cite{loshchilov2017decoupled} & AdamW~\cite{loshchilov2017decoupled} \\
    Learning rate & 1e-4 & 1e-4 & 1e-5 & 1e-5 \\
    Learning rate scheduler & Constant & Constant & Constant & Constant \\
    Weight decay & 0.05 & 0.05 & 0.05 & 0.05 \\
    Adam $\beta$ & (0.9, 0.95) & (0.9, 0.95) & (0.9, 0.95) & (0.9, 0.95) \\
    Max view num & 4 & 4 & 8 & 8 \\
    Batch size & 512 & 1024 & 1024 & 256 \\
    Steps & 100k & 200k & 30k & 30k \\
    Warmup steps & 4k & 4k & 1k & 1k \\
    Initialization & HunyuanDiT\cite{li2024hunyuan} & HunyuanDiT\cite{li2024hunyuan} & Stage 1 & Stage 2 \\
    Attention blocks (encoder) & 20 & 20 & 20 & 20 \\
    Attention blocks (decoder) & 40 & 40 & 40 & 40 \\
    \midrule
    Image resolutions & $256{\times}256$ & $256{\times}256$ & $256{\times}256$ & $512{\times}512$ \\
    Raymap resolutions & $16{\times}16$ & $16{\times}16$ & $16{\times}16$ & $32{\times}32$ \\
    Depth resolutions & $64{\times}64$ & $64{\times}64$ & $64{\times}64$ & $128{\times}128$ \\
    Datasets & Object-centric & Object-centric & All & All \\
    \bottomrule 
    \end{tabular}
    }
    \vspace{-0.3cm}
    \caption{Detailed hyper-parameters. }
    \label{tab:supp-hyperparams}
\end{table*}

\begin{table*}[]
    \centering
    \resizebox{\linewidth}{!}{
    \begin{tabular}{l|cc|cc|ccc}
    \toprule
    \multirow{2}{*}{Dataset} & \multicolumn{2}{c|}{Size} & \multicolumn{2}{c|}{Type} & \multicolumn{3}{c}{Support Modalities} \\
     & Scenes & Images & Real/Synthetic & Scene Type & RGB & Poses & Depths \\
    \midrule
    Objaverse~\cite{deitke2023objaverse} & 800K & 25M & Synthetic & Object-centric & $\checkmark$ & $\checkmark$ & Foreground only \\
    MVImageNet~\cite{yu2023mvimgnet} & 220K & 6.5M & Real & Object-centric & $\checkmark$ & $\checkmark$ & $\times$ \\
    CO3Dv2~\cite{reizenstein21co3d} & 19K & 1.5M & Real & Object-centric & $\checkmark$ & $\checkmark$ & Incomplete \\
    RealEstate10K~\cite{zhou2018stereo} & 10K & 10M & Real & Indoor/Outdoor Scene & $\checkmark$ & $\checkmark$ & $\times$ \\
    Hypersim~\cite{roberts2021hypersim} & 461 & 77K & Synthetic & Indoor Scene & $\checkmark$ & $\checkmark$ & Complete \\
    ARKitScenes~\cite{baruch2021arkitscenes} & 5K & 450K & Real & Indoor Scene & $\checkmark$ & $\times$ & Incomplete \\
    \bottomrule
    \end{tabular}
    }
    \vspace{-0.3cm}
    \caption{Dataset details.}
    \label{tab:supp-datasets}
\end{table*}

\section{Experiments}
\label{sec:supp-exp}
\subsection{DTU Dataset Split for Depth Evaluation}
In Sec.~\ref{sec:exp-depth}, we use different evaluation set for monodepth and multi-view depth evaluation. Specifically, we use the IDR \cite{yariv2020multiview} subset for monodepth because perfect foreground masks are provided, and follow previous work \cite{wang2024dust3r} to use the MVSNet \cite{yao2018mvsnet} subset for multi-view depth evaluation. 

\subsection{Point Cloud Fusion}
In Sec.~\ref{sec:exp-depth}, we back-project multi-view depth maps to point cloud. In practice, we additionally conduct geometric consistency filtering and fusion to clean the point cloud. The filtering follows \cite{yao2018mvsnet, zhang2023vis}, consisting of a combination of the following operations.
\begin{itemize}
    \item Geometric filtering. We project the pixels from the reference view to source views, find the pixel at the projection location, and project it back to reference view. Then we check the difference of the original position and the reprojected position, as well as their depths. 
    \item Geometry fusion. We project all pixels from source views to the reference views, and each pixel in the reference view may receive multiple values. We then change the original depth result to the average or the median of all the gathered values. 
\end{itemize}

\begin{table}[t]
    \centering
    \resizebox{1.0\linewidth}{!}{
    \begin{tabular}{l|cccc|cccccccc}
    \toprule
    \multirow{2}{*}{Methods} & GT & GT & GT & \multirow{2}{*}{Align} & \multicolumn{2}{c}{DTU} & \multicolumn{2}{c}{ETH3D} & \multicolumn{2}{c}{T\&T} \\
    \cline{6-11}
    & Pose & Range & Int. &  & rel $\downarrow$ & $\tau \uparrow$ & rel $\downarrow$ & $\tau \uparrow$ & rel $\downarrow$ & $\tau \uparrow$ \\
    \midrule
    DeepV2D & $\times$ & $\times$ & $\checkmark$ & med & 7.7 & 33.0 & 11.8 & 29.3 & 8.9 & 46.4 \\
    DUSt3R & $\times$ & $\times$ & $\times$ & med  & \underline{2.76} & \underline{77.32} & \textbf{4.71} & \textbf{61.74} & \textbf{5.54} & \textbf{56.38}\\
    Ours & $\times$ & $\times$ & $\checkmark$ & med & \textbf{1.85} & \textbf{85.46} & \underline{7.83} & \underline{38.80} & \underline{6.16} & \underline{49.43} \\
    \bottomrule
\end{tabular}
}
\caption{Unposed MVD evaluation on DTU, ETH3D, and T\&T.}
\label{tab:supp-multiview_depth}
\end{table}

\subsection{Ablation Study on Multi-task Training}\label{sec:supp-ablation}
In this section we compare the model trained by masked learning and the task-specific models including NVS, pose estimation and depth estimation. The latter ones have the same network architecture as the stage 1 model, but the input/output configuration of the training samples is set to only one task. All 4 models are trained from HunyuanDiT \cite{li2024hunyuan} initialization with halved batch size and total steps due to limited time and compute resources. The evaluation metrics for each task is the same as the main paper. 

\begin{figure*}
    \centering
    \resizebox{\linewidth}{!}{
    \begin{tabular}{c@{\hskip 1pt}c@{\hskip 1pt}c}
    \includegraphics[width=0.287\linewidth]{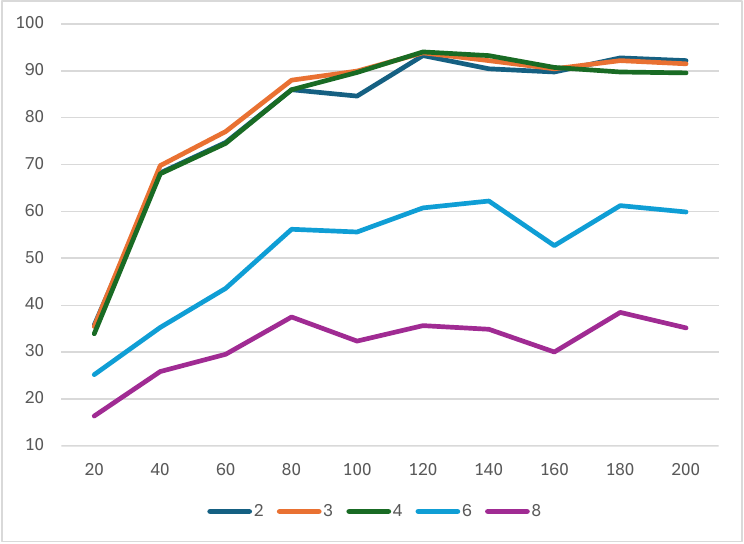} &
    \includegraphics[width=0.35\linewidth]{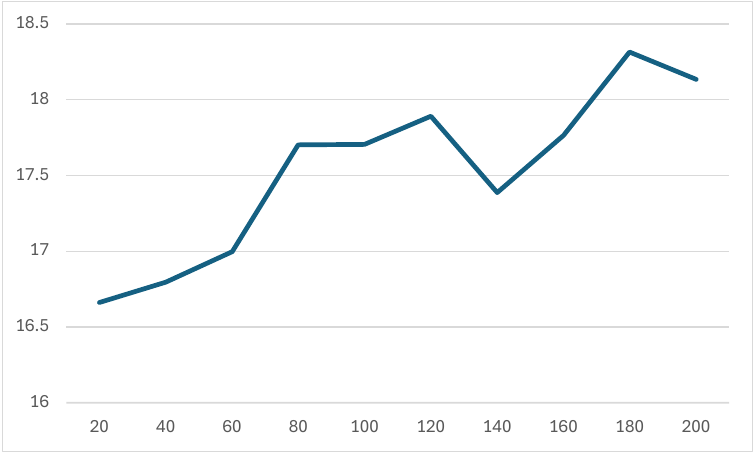} &
    \includegraphics[width=0.35\linewidth]{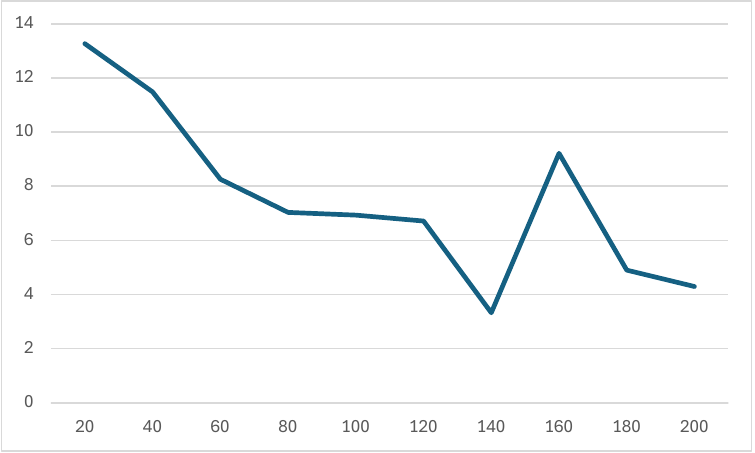} \\
    a) Pose (RRA@15$\degree\uparrow$) & b) NVS (PSNR$\uparrow$) & c) Depth (rel$\downarrow$)
    \end{tabular}}
    \caption{Evaluation results of stage 1 model for a) pose estimation, b) NVS and c) Depth estimation with respect to training step. For pose estimation we report the results for multiple view numbers. Note that the stage 1 model is only trained with view number $\leq$ 4. }
    \label{fig:supp-curve}
\end{figure*}

Quantitative reuslts are shown in Table \ref{tab:supp-abl}. The model with masked learning strategy (\textit{Multi-task}) surpasses the task-specific model in the NVS task, but fails for pose estimation and depth estimation. One possible reason is that the model capacity is shared by different tasks. Another reason related to practice is that the models for ablation studies do not fully converge. According to the evaluation curve with respect to training steps in Figure \ref{fig:supp-curve}, the model with halved batch size at 100k steps has similar performance as the full model at 60k-70k steps which still has large room of improvement. Given that our model is initialized by an RGB diffusion model, the functionality of outputing ray maps and depth maps may need longer time to converge, and thus task-specific models achieves better results within limited training time. Although not a fair comparison, note that all models for ablation study are weaker than the stage 1 and the stage 3 model. Also, in Sec.~\ref{sec:exp-hybrid} we show that the model trained by masked learning can support flexible input and boost the performance by utilizing additional input.

\begin{table*}[]
    \centering
    \resizebox{\linewidth}{!}{
    \begin{tabular}{l|ccc|ccc||l|ccc||l|cc}
    \toprule
    \multirow{2}{*}{Methods} & \multicolumn{3}{c|}{RRA @ 15$\degree\uparrow$} & \multicolumn{3}{c||}{CA @ 0.1$\uparrow$} & \multirow{2}{*}{Methods} & \multirow{2}{*}{PSNR$\uparrow$} & \multirow{2}{*}{SSIM$\uparrow$} & \multirow{2}{*}{LPIPS$\downarrow$} & \multirow{2}{*}{Methods} & \multirow{2}{*}{rel$\downarrow$} & \multirow{2}{*}{$\tau\uparrow$} \\
     & 2 & 3 & 4 & 2 & 3 & 4 &  &  &  &  &  &  &  \\
    \midrule
    Pose only & \textbf{89.2} & \textbf{86.7} & \textbf{85.8} & 100.0 & \textbf{83.1} & \textbf{77.0} & NVS only & 16.30 & 0.77 & 0.30 & Depth only & \textbf{9.07} & \textbf{26.21} \\
    Multi-task & 81.1 & 77.8 & 75.3 & 100.0 & 75.8 & 64.5 & Multi-task & \textbf{17.21} & \textbf{0.79} & \textbf{0.25} & Multi-task & 10.76 & 18.52 \\
    \midrule
    Stage 1 & 92.2 & 91.5 & 89.6 & 100.0 & 87.8 & 80.8 & Stage 1 & 18.13 & 0.81 & 0.19 & Stage 1 & 4.30 & 49.81 \\
    Stage 3 & 95.6 & 96.0 & 96.3 & 100.0 & 93.5 & 91.7 & Stage 3 & 18.87 & 0.85 & 0.21 & Stage 3 & 1.83 & 85.45 \\
    \bottomrule
    \end{tabular}
    }
    \vspace{-0.3cm}
    \caption{Ablation study on multi-task training and task-specific training. Besides different training target, the ablation models have halved batch size and total steps. The multi-task model achieves better results in NVS task but fails for pose estimation and depth estimation. One possible reason is that the multi-task model converges slower than the task-specific models. Please refer to Sec.~\ref{sec:supp-ablation} for more analysis. }
    \label{tab:supp-abl}
\end{table*}

\subsection{3D Reconstruction}
\pseudopara{Camera trajectory generation} We build different camera trajectories for generating novel views depending on different reconstruction tasks. For monocular image input, we create an orbital trajectory and sample 80 cameras evenly. All cameras are set as look-at to the world center. For sparse-view image input, we fitted a spline trajectory from the input poses, and scaled up the trajectories two times, resulting in $240 (=80 \times 3)$ views.

\pseudopara{3DGS optimization} The proposed 3DGS optimization system in built upon the open-source pipeline~\cite{nerfstudio} with several modifications. For each optimization step, we optimize Gaussian points on mini-batch images instead of single images. Besides of original L1 loss and SSIM loss, we adopt additional losses to improve the reconstruction robustness, including LPIPS loss $L_{\textrm{LPIPS}}$~\cite{zhang2018perceptual}, mask loss $L_{\textrm{mask}}$, accumulation regularization $L_{\textrm{accum}}$, absolute depth loss $L_{\textrm{depth}}$, and relative depth ranking loss $L_{\textrm{rel-depth}}$~\cite{wang2023sparsenerf}. The accumulation regularization is designed to constrain the alpha values of Gaussian points to be either fully opaque or completely transparent, aiming to reduce floaters in the scene. It is composed of a binary cross-entropy loss and entropy loss:
\begin{equation*}
    L_{\textrm{accum}} = BCE(\alpha, 0.5) -\alpha \log(\alpha) + (1-\alpha) \log(1-\alpha),
\end{equation*}
where $\alpha$ denotes the accumulation values.

For monocular 3D reconstruction, the value for each loss is set to $w_{\textrm{L1}}=1.0, w_{\textrm{SSIM}}=0.2, w_{\textrm{LPIPS}}=10.0, w_{\textrm{mask}}=5.0, w_{\textrm{accum}}=5.0$. Depth loss is not applied in the optimization. The mini-batch size for each step is set to $10$. For the input view, the weight of L1 loss is specifically set to $10.0$ for high significance. 

For sparse-view 3D reconstruction, the weight values are set to $w_{\textrm{L1}}=1.0, w_{\textrm{SSIM}}=0.2, w_{\textrm{LPIPS}}=10.0, w_{\textrm{mask}}=5.0, w_{\textrm{accum}}=0.5, w_{\textrm{depth}}=10.0, w_{\textrm{rel-depth}}=20.0$. The mini-batch size for each step is set to $5$, and the L1 loss weight of input views is set to $20$. 

We use the back-projected point cloud as Gaussian point initialization. Similar to CAT3D~\cite{gao2024cat3d}, we conduct in total of 1200 and 3000 optimization steps for two tasks, respectively. We apply the scale regularization~\cite{xie2024physgaussian} to constrain the extreme Gaussian scales.

\subsection{Limitation}

During experiments, we found that our model performs well on object-centric and indoor scenes but degrades in outdoor environments, primarily due to the lack of large-scale outdoor training data—our dataset consists of objects and limited indoor scenes. The lack of high-quality outdoor data is a common issue in the community, and similar problem has been noticed in other models.

\cref{tab:supp-multiview_depth} demonstrates unposed depth prediction results on ETH3D and T\&T. Our model performs worse than DUSt3R (trained on outdoor datasets), but still surpass DeepV2D.

\begin{figure*}[t]
    \centering
    \includegraphics[width=1.0\linewidth]{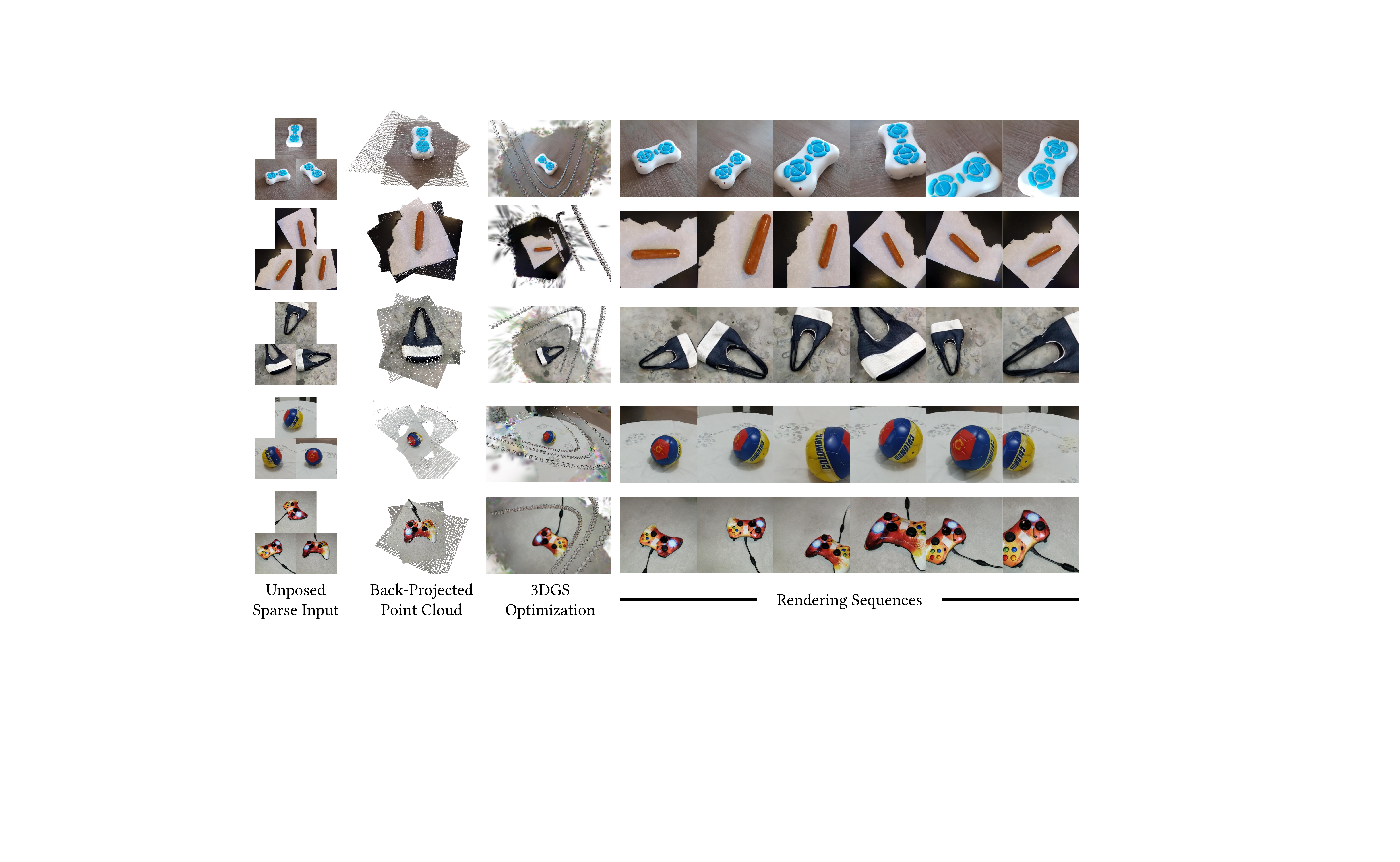}
    \vspace{-6mm}
    \caption{More unposed sparse-view 3D reconstruction results. }
    \vspace{-4mm}
    \label{fig:supp-unposed-vis}
\end{figure*}

\begin{figure*}[t]
    \centering
    \includegraphics[width=1.0\linewidth]{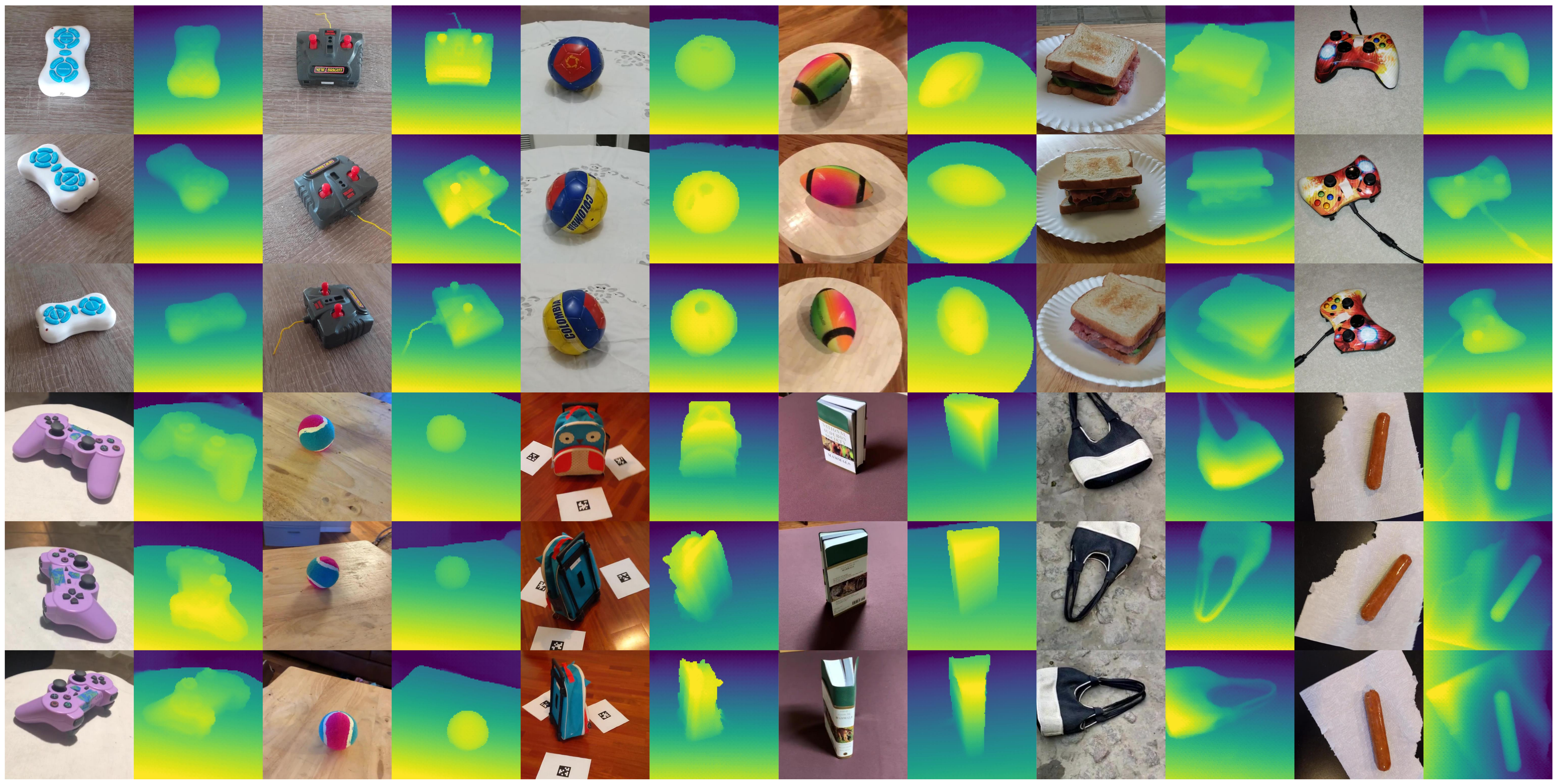}
    \vspace{-6mm}
    \caption{Visualization of multi-view depth prediction results. }
    \vspace{-4mm}
    \label{fig:supp-depth-vis}
\end{figure*}

\subsection{More visualization}
Here we present more visualization results about unposed sparse-view 3D reconstruction (Fig.~\ref{fig:supp-unposed-vis}) and multi-view depth predictions (Fig.~\ref{fig:supp-depth-vis}).

\end{document}